\def\year{2019}\relax
\documentclass[letterpaper]{article} 
\usepackage{aaai19}  
\usepackage{times}  
\usepackage{helvet}  
\usepackage{courier}  
\usepackage{url}  
\usepackage{graphicx}  
\title{Spatial Mixture Models with Learnable Deep Priors for Perceptual Grouping}
\author{Jinyang Yuan, Bin Li\protect\thanks{Corresponding author}, Xiangyang Xue\\
Shanghai Key Laboratory of Intelligent Information Processing\\
School of Computer Science, Fudan University\\
Fudan-Qiniu Joint Laboratory for Deep Learning\\
Shanghai Institute of Intelligent Electronics \& Systems\\
\{yuanjinyang, libin, xyxue\}@fudan.edu.cn\\
}

\usepackage{amsmath}
\usepackage{algorithm}
\usepackage{algpseudocode}
\usepackage{siunitx}
\usepackage{subfig}
\usepackage{multirow}
\DeclareMathOperator*{\argmax}{arg\,max}
\newcolumntype{C}[1]{>{\centering\let\newline\\\arraybackslash\hspace{0pt}}m{#1}}
\newcommand{\STAB}[1]{\begin{tabular}{@{}c@{}}#1\end{tabular}}

\usepackage{color}

\begin{document}

\maketitle

\begin{abstract}
Humans perceive the seemingly chaotic world in a structured and compositional way with the prerequisite of being able to segregate conceptual entities from the complex visual scenes. The mechanism of grouping basic visual elements of scenes into conceptual entities is termed as perceptual grouping. In this work, we propose a new type of spatial mixture models with learnable priors for perceptual grouping. Different from existing methods, the proposed method disentangles the attributes of an object into ``shape'' and ``appearance'' which are modeled separately by the mixture weights and the mixture components. More specifically, each object in the visual scene is fully characterized by one latent representation, which is in turn transformed into parameters of the mixture weight and the mixture component by two neural networks. The mixture weights focus on modeling spatial dependencies (i.e., shape) and the mixture components deal with intra-object variations (i.e., appearance). In addition, the background is separately modeled as a special component complementary to the foreground objects. Our extensive empirical tests on two perceptual grouping datasets demonstrate that the proposed method outperforms the state-of-the-art methods under most experimental configurations. The learned conceptual entities are generalizable to novel visual scenes and insensitive to the diversity of objects. Code is available at {\color{blue}\url{https://github.com/jinyangyuan/learnable-deep-priors}.}
\end{abstract}

\section{Introduction}

The ability to perceive complex visual scenes in a structured and compositional way is crucial for humans to understand the seemingly chaotic world \cite{lake2017building}. Finding the underlying mechanisms of grouping basic visual elements is usually termed as the binding \cite{treisman1996binding,wolfe1999psychophysical}, or more specifically, the perceptual grouping \cite{grossberg1997visual} problem, and has been studied extensively in the fields of neuroscience, cognitive science and psychology. Designing more human-like AI systems which learn compositional and disentangled representations \cite{bengio2013representation} is desirable, because of their superior expressiveness and generalization ability compared to those systems learning a single complex representation. Complex visual scenes are composed of numerous relatively simpler primitive entities. By hierarchically decomposing the observed visual scenes \cite{bienenstock1997compositionality}, the raw observations can be summarized into organized and compact knowledge which is generalizable to an infinite number of novel scenes constructed by the combinations of these primitive entities \cite{biederman1987recognition,hummel1992dynamic,van2015part}.

Inspired by the synchronization theory \cite{milner1974model}, various approaches have been proposed to tackle the perceptual grouping problem by using neuronal synchrony as the grouping mechanism \cite{wang1995locally,rao2008unsupervised,reichert2013neuronal}. Despite of the  satisfactory grouping results already obtained, these methods do not learn separate representations for individual objects in the visual scene, which limits the expressiveness and generalization ability of the learned models. In order to learn disentangled and compositional representations of visual scenes, several methods which integrate spatial mixture models with neural networks have been proposed in recent years \cite{greff2015binding,greff2016tagger,premont2017recurrent,greff2017neural}. In these approaches, different objects are model by different mixture components, and each component is encouraged to focus on an individual object. The underlying intuition is that the image of visual scene can be decomposed into the pixel-wise weighted sum (modeled by mixture weights) of reconstructed images of individual objects (modeled by mixture components).

The perceptual grouping results of these recently proposed spatial mixture models are gratifying. Images of individual objects can be reconstructed accurately if objects do not overlap each other. However, there exist two potential downsides: 1) different attributes (e.g. shape, appearance) of objects cannot be well disentangled, and 2) the background is either not explicitly modeled by a mixture component or considered as an ordinary component, which in turn either complicates the determination of background region or increases the modeling complexity because the background region is much harder to represent due to the diverse combinations of foreground objects.

In this paper, we propose a new type of spatial mixture models which places a learnable deep prior on the model parameters and considers modeling the background as a special component complementary to the foreground objects. Parameters of the learnable deep prior consist of both mixture weight parameters and mixture component parameters, and are computed based on the outputs of two \emph{learnable} neural networks with latent representations of objects as inputs. The basic idea is that, in the proposed model, the mixture weights should focus on modeling the spatial dependencies (i.e., shapes) of the objects while the mixture components deal with variations (i.e., appearance) within objects. The inherent structure and learned parameters of the network regularize the model to correctly disentangle and estimate the ``shapes'' and ``appearances'' of objects in the visual scenes. By constructing the mixture weights in an innovative manner inspired by the truncated stick-breaking process, foreground objects and the background which is harder to represent are modeled differently. Because the background is considered as a special component whose mixture weights are implicitly determined by estimations of less varying foreground objects, the proposed method can thus achieve better compositionality and generalizes better to novel visual scenes.

We evaluate our method on two perceptual grouping datasets, in which images are composed of simple shapes or handwritten images, under different experimental configurations. Extensive empirical results suggest that representing the complex regions of background pixels in a compositional manner is crucial to high-quality grouping results. Compared with two state-of-the-art perceptual grouping algorithms, the proposed method not only achieves comparable or higher grouping accuracies, but also better estimates individual objects from the observed visual scene.
The learned conceptual entities are generalizable to novel visual scenes and insensitive to the diversity of objects.

\section{Preliminaries}

\subsection{Perceptual Grouping}

Perceptual grouping is a type of binding problem that studies how brains arrange elements of visual scenes into conceptual entities like objects or patterns. Based on the inductive biases that are either innate or learned from experiences, humans tend to segregate, for example, the upper-left corner image in Figure \ref{fig:shapes_tagger} into three white hollow shapes and one black background although these shapes are connected and partially overlapped. The underlying mechanisms have been studied extensively for years. Gestalt psychologists suggested that humans perceive meaningful contents from the sensory information based on inborn mental laws, and summarized their theories into several rule-based principles of perception (e.g. proximity, similarity, continuity, closure, and common fate) which is often known as Gestalt laws of grouping \cite{goldstein2016sensation}. The principle of common fate differs from others in that it requires temporal information and is only applicable when relative motions exist. In this work, we only consider utilization of spatial information and focus on building models which follow the principles of proximity, similarity, continuity, and closure by learning prior knowledge of spatial dependencies from data in an unsupervised setting.

\subsection{Spatial Mixture Model}

In spatial mixture models, pixels are assumed to be generated independently from the mixture. Correlations between pixels can be achieved by assuming mixture components, or mixture weights, or both of them to be spatially dependent. Let $X_m$ be the $m$th pixel of image $\boldsymbol{X}$, and $Z_m$ be the latent variable specifying the mixture component of $X_m$. $\boldsymbol{\theta}_{m,k} = \{\boldsymbol{\theta}_{m,k}^c, \boldsymbol{\theta}_{m,k}^w\}$ represents parameters for generating $X_m$ in the $k$th component. For the sake of notational simplicity, we let $p_{m,k}=P(X_m|Z_m\!\!=\!k; \boldsymbol{\theta}_{m}^c)$ and $\pi_{m,k}=P(Z_m\!\!=\!k; \boldsymbol{\theta}_{m}^w)=\theta_{m,k}^w$. The log probability of observing image $\boldsymbol{X}$ is given by
\begin{equation}
	\label{equ:spatial_mixture_model}
	\log{P(\boldsymbol{X}; \boldsymbol{\theta})} = \sum_{m}{\log{\sum_{k}{p_{m,k} \pi_{m,k}}}}
\end{equation}

Computing the maximum likelihood estimate (MLE) of $\boldsymbol{\theta}$ directly is difficult due to the existence of unobserved component assignments. Approximated estimates can be obtained iteratively by applying the Expectation-Maximization (EM) algorithm. Let $\gamma_{m,k}^{(t)}$ denote the posterior probability at the $t$th iteration $P(Z_m\!\!=\!k|X_m; \boldsymbol{\theta}_{m}^{(t)})$, which is computed by 
\begin{equation}
	\label{equ:posterior}
	\gamma_{m,k}^{(t)} = \frac{p_{m,k}^{(t)} \pi_{m,k}^{(t)}}{\sum_{k'}{p_{m,k'}^{(t)} \pi_{m,k'}^{(t)}}}
\end{equation}
Parameters are updated using $\boldsymbol{\theta}^{(t+1)} \!=\! \argmax_{\boldsymbol{\theta}}{Q(\boldsymbol{\theta}; \boldsymbol{\theta}^{(t)})}$, where $Q(\boldsymbol{\theta}; \boldsymbol{\theta}^{(t)})$ is the Q-function defined as
\begin{equation}
	\label{equ:q_function}
	Q(\boldsymbol{\theta}; \boldsymbol{\theta}^{(t)}) = \sum_{m}{\sum_{k}{\gamma_{m,k}^{(t)}(\log{p_{m,k}} + \log{\pi_{m,k}})}}
\end{equation}

\subsection{Neural Expectation Maximization}

Neural Expectation Maximization (N-EM) \cite{greff2017neural} is a state-of-the-art spatial mixture model for solving the perceptual grouping problem. In N-EM, mixture components and mixture weights are modeled as spatially dependent and spatially independent, respectively. Parameters $\boldsymbol{\theta}_{m,k}^c \!\!=\!\! f_{\phi}(\boldsymbol{s}_k)_m$ are generated by a neural network $f_{\phi}$ with latent representations of objects $\boldsymbol{s}_k$ as inputs, and mixture weights $\pi_{m,k}$ which are identical at all pixels are either computed based on the EM update rule or fixed to $1 / K$ ($K$ is the number of mixture components) for simplicity \cite{greff2015binding}.

Because of the non-linearity of neural networks, the optimal latent representation $\boldsymbol{s}_k^*$ that maximizes $Q(\boldsymbol{\theta}; \boldsymbol{\theta}^{(t)})$ does not have a closed-form solution. Two types of generalized EM algorithms were proposed to solve this problem. One is to utilize the gradient descent to optimize $\boldsymbol{s}_k$ iteratively
\begin{equation}
	\boldsymbol{s}_k^{(t+1)} = \boldsymbol{s}_k^{(t)} + \eta \sum_{m}{\gamma_{m,k}^{(t)} \frac{\partial \log{p_{m,k}}}{\partial \boldsymbol{\theta}_{m,k}^c} \frac{\partial \boldsymbol{\theta}_{m,k}^c}{\partial \boldsymbol{s}_k}}\bigg|_{\boldsymbol{s}_k = \boldsymbol{s}_k^{(t)}}
\end{equation}
The other is to mimic the process of gradient descent by a recurrent neural network (RNN). The neural network $f_{\phi}$ is trained by the backpropagation through time (BPTT) algorithm \cite{robinson1987utility,werbos1988generalization}.

The loss function of each image is given by
\begin{align}
	\label{equ:loss_nem}
	L = & -\sum_{m}{\sum_{k}{\gamma_{m,k} (\log{p_{m,k}} + \log{\pi_{m,k}})}} \nonumber \\
	& + \lambda \sum_{m}{\sum_{k}{(1 - \gamma_{m,k}) D_{\text{KL}}(p_{\text{prior}}||p_{m,k})}}
\end{align}
This loss function consists of two parts. The first part is the negative Q-function which measures the reconstruction errors. The second part is the weighted sum of the Kullback-Leibler (KL) divergences between the mixture components $p_{m,k}$ and a distribution $p_{\text{prior}}$ predefined by the prior knowledge of the image (e.g. the background intensities). $p_{m,k}$ is encouraged to be close to the distribution of background when the $k$th component is unconfident that the $m$th pixel should be assigned to it ($\gamma_{m,k}$ is small).

\section{Spatial Mixture Model with\\Learnable Deep Priors}

In N-EM, only spatially dependent mixture components are transformed from latent representations of objects. Spatially independent mixture weights are modeled as coefficients which measure the importances of mixture components in the mixture model.
As a result, different attributes are entangled in mixture components and the region occupied by each object cannot be determined solely based on the latent representation. Moreover, the background is not explicitly modeled by a mixture component, and cannot be determined via the Maximum \emph{a posteriori} (MAP) estimation.

To better represent individual objects and model the background pixels, we integrate neural networks and spatial mixture models in a different way.
Both mixture weights and mixture components are transformed from latent representations that fully characterize attributes of objects.
Attributes are disentangled into ``shapes'' and ``appearances'', and modeled separately by mixture weights and mixture components. To follow the principle of similarity in the Gestalt laws of grouping, all pixels share the same mixture component parameters. In the following, we first introduce our approach in a general framework named Learnable Deep Priors and then make a concrete example for clarity.

\subsection{Learnable Deep Priors}

Let $P(\boldsymbol{\theta}_m'; \boldsymbol{\theta}_m) \!=\! \sum_k{\pi_{m,k} \delta(\boldsymbol{\theta}_m' - \boldsymbol{\theta}_{m,k}^c)}$, where $\delta(\cdot)$ is the Dirac delta function. The log probability \eqref{equ:spatial_mixture_model} of the observed image $\boldsymbol{X}$ can be rewritten as
\begin{equation*}
	\log{P(\boldsymbol{X}; \boldsymbol{\theta})} = \sum_{m}{\log{\int{P(X_m|\boldsymbol{\theta}_m') P(\boldsymbol{\theta}_m'; \boldsymbol{\theta}_m) d\boldsymbol{\theta}_m'}}}
\end{equation*}
where the prior $P(\boldsymbol{\theta}_m'; \boldsymbol{\theta}_m)$ of model parameters $\boldsymbol{\theta}_m'$ is defined with both \emph{mixture weights} $\pi_{m,k}$ and \emph{mixture component parameters} $\boldsymbol{\theta}_{m,k}^c$. Because $\pi_{m,k}$ and $\boldsymbol{\theta}_{m,k}^c$ are transformed from latent representations of objects by \emph{learnable} neural networks $f_{\phi}$ and $g_{\psi}$, $P(\boldsymbol{\theta}_m'; \boldsymbol{\theta}_m)$ is referred to as the \emph{learnable deep prior} in our method.

\textbf{Truncated Stick-Breaking Mixture Weights}.
The spatially dependent mixture weights $\pi_{m,k}$ are computed based on outputs of a neural network $f_{\phi}$ and are constrained by the structure and parameters of the network. Let $K$ be the number of components, $\boldsymbol{s}_k$ the latent representation of the $k$th component, and $\sigma(\cdot)$ the sigmoid function. $c_{k,m}$ is defined as $\sigma(f_{\phi}(\boldsymbol{s}_k)_m)$. The computation of $\pi_{m,k}$ which shares a similar form to the truncated stick-breaking process gives
\begin{equation}
	\label{equ:prior}
	\pi_{m,k} =
	\begin{cases}
		c_{k,m} \prod_{k'<k}{(1 - c_{k',m})}, & k < K \\
		\prod_{k'<K}{(1 - c_{k',m})}, & k = K
	\end{cases}
\end{equation}
\eqref{equ:prior} differs from the truncated stick-breaking process in that the length of stick $c_{k,m}$ is computed by a neural network instead of sampled from the beta distribution. Only the first $K - 1$ components which model foreground objects are assigned with latent representations. The complex-shaped background region is estimated based on the predicted shapes of these foreground objects.

Applying the softmax function to the outputs of neural networks is another viable way of modeling mixture weights. The main consideration of choosing \eqref{equ:prior} over the softmax function is \emph{compositionality}. The purpose of applying the mixture model is to decompose the complex visual scene into simpler components. Generally speaking, objects in the visual scene are relatively easy to represent. The background pixels, however, are almost as hard to model as the visual scene due to the diverse combinations of foreground objects. Because the softmax function treats each component equally, the complex-shaped background region need to be represented directly and the modeling complexity is increased significantly.

\textbf{Mixture Component Parameters}. 
To follow the principle of similarity in the Gestalt laws of grouping, mixture component parameters $\boldsymbol{\theta}_{m,k}^c$ are modeled as spatially independent. If no additional constraint is added to $\boldsymbol{\theta}_{m,k}^c$, the optimal solution that maximizes the Q-function can be obtained by setting $\sum_{m}{\partial Q(\boldsymbol{\theta}; \boldsymbol{\theta}^{(t)}) / \partial \boldsymbol{\theta}_{m,k}^c}$ to zero. To fully characterize attributes of objects with latent representations, all $\boldsymbol{\theta}_{m,k}^c$ with $k < K$ are outputs of a neural network $g_{\psi}$ with latent representations $\boldsymbol{s}_k$ as inputs and are optimized by updating $\boldsymbol{s}_k$. Parameters $\boldsymbol{\theta}_{m,K}^c$ for the background component are modeled as learnable variables that are optimized by gradient descent in the M-step of the EM algorithm.

\subsection{Optimization of Mixture Model Parameters}

Latent representations $\boldsymbol{s}_{k}$ can be updated by either gradient descent with a learnable learning rate or an RNN which imitate the behavior of gradient descent. For the sake of notational simplicity, outputs of the neural networks $f_{\phi}(\boldsymbol{s}_k)_m$ and $g_{\psi}(\boldsymbol{s}_k)$ are denoted by $f_{k,m}$ and $g_k$, respectively. If using gradient descent as the update rule, latent representations $\boldsymbol{s}_k$ with $1 \leq k < K$ are updated by
\begin{multline}
	\label{equ:update_proposed}
	\boldsymbol{s}_k^{(t+1)} = \boldsymbol{s}_k^{(t)} + \eta_s \sum_{m}\Big(\gamma_{m,k}^{(t)} \frac{\partial \log{p_{m,k}}}{\partial g_k} \frac{\partial g_k}{\partial \boldsymbol{s}_k} \\
	+ \sum_{k' \geq k}{\gamma_{m,k'}^{(t)} \frac{\partial \log{\pi_{m,k'}}}{\partial f_{k,m}} \frac{\partial f_{k,m}}{\partial \boldsymbol{s}_k}}\Big)\bigg|_{\boldsymbol{s}_k = \boldsymbol{s}_k^{(t)}}
\end{multline}

If applying an RNN to learn the procedure of gradient descent, inputs to the RNN are features extracted from the concatenation of $\sum_{m}{\gamma_{m,k}^{(t)} (\partial \log{p_{m,k}} / \partial g_k)}$ and all $\sum_{k' \geq k}{\gamma_{m,k'}^{(t)} (\partial \log{\pi_{m,k'}} / \partial f_{k,m})}$ with $1 \leq m \leq M$ by an encoder network.

The update rule for the parameters $\boldsymbol{\theta}_{m,K}^c$ is given by
\begin{equation}
	\label{equ:update_last}
	{\boldsymbol{\theta}_{m,K}^{c^{(t+1)}}} \!\!= {\boldsymbol{\theta}_{m,K}^{c^{(t)}}} + \eta_{\theta} \sum_{m'}{\gamma_{m'\!,K}^{(t)} \frac{\partial \log{p_{m'\!,K}}}{\partial \boldsymbol{\theta}_{m'\!,K}^c}}\bigg|_{\boldsymbol{\theta}_{m'\!\!,K}^c = {\boldsymbol{\theta}_{m'\!\!,K}^{c^{(t)}}}}
\end{equation}

\subsection{Learning of Neural Networks}

Neural networks $f_{\phi}$ and $g_{\psi}$ are trained by BPTT algorithm. The loss function is chosen as
\begin{align}
	\label{equ:loss_proposed}
	L = & -\sum_{m}{\sum_{k}{\gamma_{m,k} (\log{p_{m,k}} + \log{\pi_{m,k}})}} \nonumber \\
	& + \lambda \sum_{m}{D_{\text{KL}}(p_{\text{prior}}||p_{m,K})}
\end{align}
As with the loss function \eqref{equ:loss_nem} used in N-EM, \eqref{equ:loss_proposed} also consists of the Q-function term and a regularization term. This regularization acts like a simplified version of the inductive biases required for human to solve the figure-ground organization problem (distinguishing a figure from the background). In N-EM, prior information of the background $p_{\text{prior}}$ is used to regularize all mixture components. In our method, the last mixture component $p_{m,K}$ which models the background is encouraged to be similar to $p_{\text{prior}}$, so that the network can better focus on modeling the foreground objects.

\subsection{Example: Gaussian Mixture Components}

The conditional probability distribution of a pixel given its component assignment is usually assumed to be Gaussian for real-valued images, and Bernoulli for binary images. We consider only Gaussian distribution in this section (detailed procedures for each image in an epoch are described in Algorithm \ref{alg:proposed}). Formulas for other types of probability distributions can be derived with little modification. Only the means of the Gaussian distributions are learnable parameters, and the variances are assumed to be constant and set to $\alpha^{-1}$ with $\alpha$ being a hyperparameter tuned by cross-validation. Except for the last component, the means of the Gaussian distributions are computed by the neural network $g_{\psi}$. For the $K$th component, it is modeled as a parameter $\mu_K$ that is learned by gradient descent. Let $x_m$ be the intensity of the $m$th pixel. The mixture component $p_{m,k}$ is computed by
\begin{equation}
	\label{equ:conditional_gaussian}
	p_{m,k} =
	\begin{cases}
		\mathcal{N}(x_m; g_{\psi}(\boldsymbol{s}_k), \alpha^{-1}), & k < K \\
		\mathcal{N}(x_m; \mu_K, \alpha^{-1}), & k = K
	\end{cases}
\end{equation}

\begin{algorithm}[t]
	\caption{Proposed Method (Gaussian Distribution)}
	\label{alg:proposed}
	\begin{algorithmic}
		\State // Initialization:
		\State Draw $\boldsymbol{s}_k^{(0)}$ randomly from $\mathcal{N}(\boldsymbol{0}, \sigma_{\text{init}}^2 \boldsymbol{I})$; \quad $\mu_K^{(0)} \gets \mu_{\text{prior}}$;
		\For{$t \gets 0$ to $T-1$}
		\State // E-Step:
		\State $f_{k,m}^{(t)} \gets f_{\phi}(\boldsymbol{s}_k^{(t)})_m$; \quad $g_k^{(t)} \gets g_{\psi}(\boldsymbol{s}_k^{(t)})$;
		\State Compute $p_{m,k}^{(t)}$, $\pi_{m,k}^{(t)}$ and $\gamma_{m,k}^{(t)}$ using \eqref{equ:conditional_gaussian}, \eqref{equ:prior} and \eqref{equ:posterior};
		\State // M-Step:
		\If{using gradient descent}
		\State Update $\boldsymbol{s}_k^{(t+1)}$ using \eqref{equ:update_proposed_gaussian};
		\Else
		\State $u \gets \alpha \sum_{m}{\gamma_{m,k}^{(t)} (x_m - g_k^{(t)})}$;
		\State $v_m \gets \gamma_{m,k}^{(t)} - \sum_{k' \geq k}{\gamma_{m,k'}^{(t)}} \sigma(f_{k,m}^{(t)})$;
		\State $r_{\text{in}} \gets enc([u, v_1, v_2, \dots, v_M])$;
		\State $\boldsymbol{s}_k^{(t+1)} \gets \text{RNN}(r_{\text{in}}, \boldsymbol{s}_k^{(t)})$;
		\EndIf
		\State Update $\mu_K^{(t+1)}$ using
		\eqref{equ:update_last_gaussian};
		\State // Loss:
		\State Compute $L^{(t)}$ using \eqref{equ:loss_proposed};
		\EndFor
		\State // Neural Networks and Learning Rates:
		\State Compute the weighted sum of losses $L_{\text{sum}} \gets \sum_{t}{w_t L^{(t)}}$;
		\State Update $\phi$, $\psi$, $\eta_s$ and $\eta_{\theta}$ based on $L_{\text{sum}}$;
	\end{algorithmic}
\end{algorithm}

Substituting \eqref{equ:conditional_gaussian} and \eqref{equ:prior} into \eqref{equ:update_proposed}, the gradient descent update rule for latent representations becomes
\begin{multline}
	\label{equ:update_proposed_gaussian}
	\boldsymbol{s}_k^{(t+1)} = \boldsymbol{s}_k^{(t)} + \eta_s \sum_{m}\Big(\alpha \gamma_{m,k}^{(t)} (x_m - g_k^{(t)}) \frac{\partial g_k}{\partial \boldsymbol{s}_k} \\
	+ \big(\gamma_{m,k}^{(t)} - \sum_{k' \geq k}{\gamma_{m,k'}^{(t)}} \sigma(f_{k,m}^{(t)})\big) \frac{\partial f_{k,m}}{\partial \boldsymbol{s}_k}\Big)\bigg|_{\boldsymbol{s}_k = \boldsymbol{s}_k^{(t)}}
\end{multline}

If updating $\boldsymbol{s}_k$ with an RNN, the encoder network $enc$ transforms the concatenation of $\alpha \sum_{m}{\gamma_{m,k}^{(t)} (x_m - g_k^{(t)})}$ and $\gamma_{m,k}^{(t)} - \sum_{k' \geq k}{\gamma_{m,k'}^{(t)}} \sigma(f_{k,m}^{(t)}), 1 \leq m \leq M$ to features $r_{\text{in}}$ as the inputs to the RNN.

Substituting \eqref{equ:conditional_gaussian} into \eqref{equ:update_last} and replacing $\boldsymbol{\theta}_{m,K}^c$ with $\mu_K$, the learnable parameter for the last component is updated by
\begin{equation}
	\label{equ:update_last_gaussian}
	\mu_K^{(t+1)} = \mu_K^{(t)} + \eta_{\theta} \alpha \sum_{m}{\gamma_{m,K}^{(t)} (x_m - \mu_K^{(t)})}
\end{equation}

\section{Related Work}

Several approaches have been proposed to solve the perceptual grouping problem and related tasks in recent years. Tagger \cite{greff2016tagger} combines the iterative amortized grouping (TAG) mechanism and the Ladder Network \cite{rasmus2015semi} to learn perceptual grouping in an unsupervised manner. It utilizes multiple copies of the same neural network to model different groups in the visual scene and iteratively refine the reconstruction result. RTagger \cite{premont2017recurrent} replaces the Ladder Network with the Recurrent Ladder Network and extends Tagger to sequential data. Neural Expectation Maximization (N-EM) \cite{greff2017neural} tackles the problem based on the Expectation-Maximization (EM) framework \cite{dempster1977maximum}, and achieves comparable performance to Tagger with much fewer parameters. Relational Neural Expectation Maximization (R-NEM) \cite{van2018relational} integrates N-EM with a type of Message Passing Neural Network \cite{gilmer2017neural} to learn common-sense physical reasoning based on the compositional object-representations extracted by N-EM.

Our method is mostly related to N-EM. Same as N-EM, the update rules for the distributed representations of objects in the visual scene are derived from the EM algorithm. Different from N-EM, the foreground objects and background are modeled separately, and attributes of objects are disentangled into ``shapes'' and ``appearances'' and modeled by mixture weights and mixture components, respectively.

\section{Experiments}

\subsubsection{Datasets}
Our method is evaluated on two datasets derived from a set of publicly released perceptual grouping datasets provided by \cite{greff2015binding,greff2016tagger,greff2017neural}. We refer to these two datasets as the \textit{Multi-Shapes} dataset and the \textit{Multi-MNIST} dataset. The Multi-Shapes dataset consists of $4$ subsets, which differs from each other in either the image size ($20 \!\times\! 20$ or $28 \!\times\! 28$) or the number of objects in each image ($2$, $3$ or $4$). Each subset contains $\num[group-separator={,}]{70000}$ binary images with simple shapes located at random positions. The Multi-MNIST dataset is composed of $3$ subsets with different degrees of object variations. In each subset, there are $\num[group-separator={,}]{70000}$ grayscale $48 \!\times\! 48$ images which contain $2$ handwritten digits. Images in both Multi-Shapes and Multi-MNIST datasets may contain overlapped objects.

\subsubsection{Comparison Methods}
The proposed method is compared with two state-of-the-art perceptual grouping algorithms, namely, \textit{N-EM} \cite{greff2017neural} and \textit{Tagger} \cite{greff2016tagger}. To assess the effectiveness of computing mixture weights using \eqref{equ:prior}, we substitute it with the softmax function and evaluate this modification (refered as \textit{softmax method}) on the Multi-Shapes dataset. Neural networks are trained with the Adam optimization algorithm \cite{kingma2014adam} for all approaches. To make a fair comparison, we do not specially design the neural networks $f_{\phi}$ and $g_{\psi}$ for the proposed method. The structure of $f_{\phi}$ which generates mixture weights in our method is identical to the network $f_{\phi}$ in N-EM. The extra $g_{\psi}$ for computing mixture component parameters is simply a one-layer fully-connected neural network, in which the number of parameters equals the dimension of latent representations and negligible compared to the total number of network parameters.

\subsubsection{Evaluation Metrics}
We use two metrics to evaluate the grouping results. The first is \textit{Adjusted Mutual Information} (AMI), which measures the accuracy of mixture assignments. To be consistent with previous work \cite{greff2015binding,greff2016tagger,greff2017neural}, AMI scores are not computed in the background and overlapping regions. The second is \textit{Mean Squared Error} (MSE), which computes the smallest mean squared differences between the estimated images of individual objects and the ground truth images over all possible permutations (formulated as the assignment problem and solved efficiently by the Hungarian method). MSE scores are evaluated at all pixels.

\subsection{Perceptual Grouping Performance Comparisons}

\begin{figure}[!ht]
	\centering
	\subfloat[Tagger (Size $20 \times 20$)]{\includegraphics[width=\linewidth]{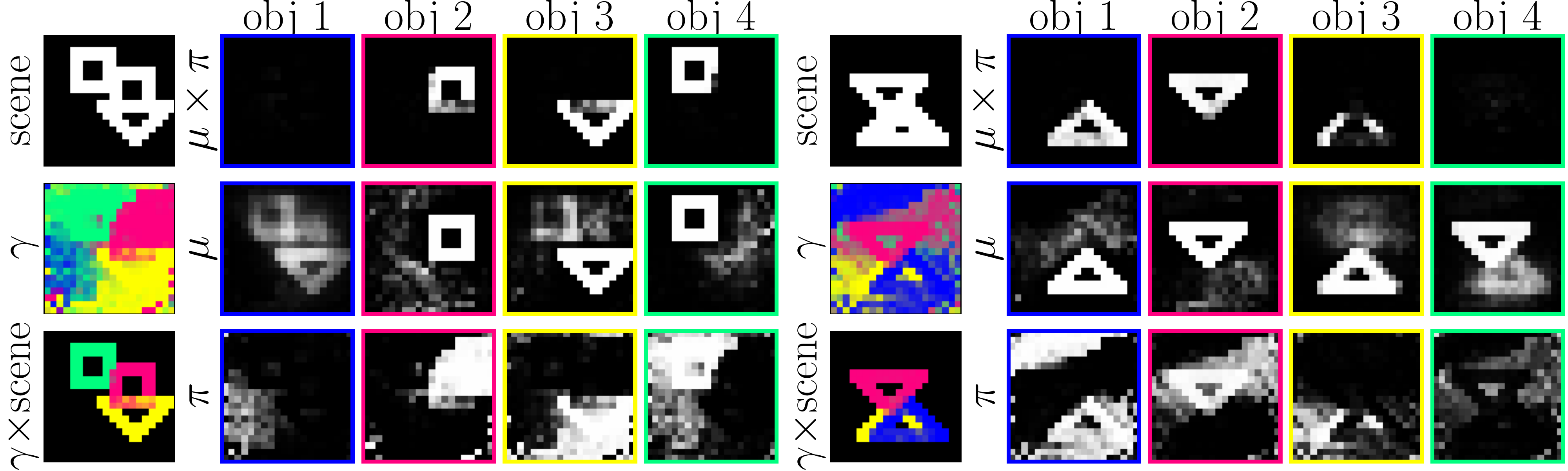}%
		\label{fig:shapes_tagger}}
	\\
	\subfloat[N-EM (Size $28 \times 28$)]{\includegraphics[width=\linewidth]{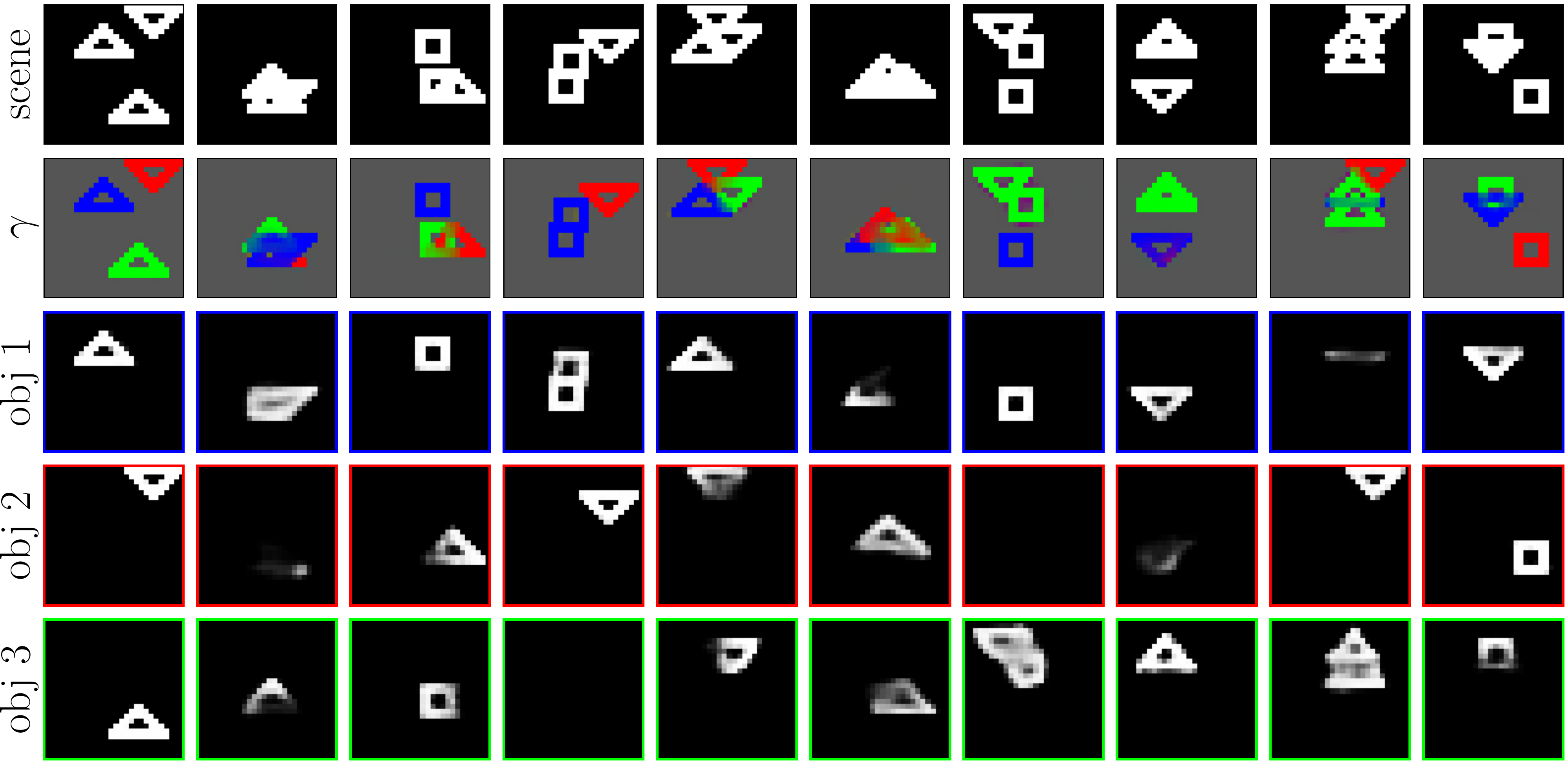}%
		\label{fig:shapes_nem}}
	\\
	\subfloat[Proposed (Size $28 \times 28$)]{\includegraphics[width=\linewidth]{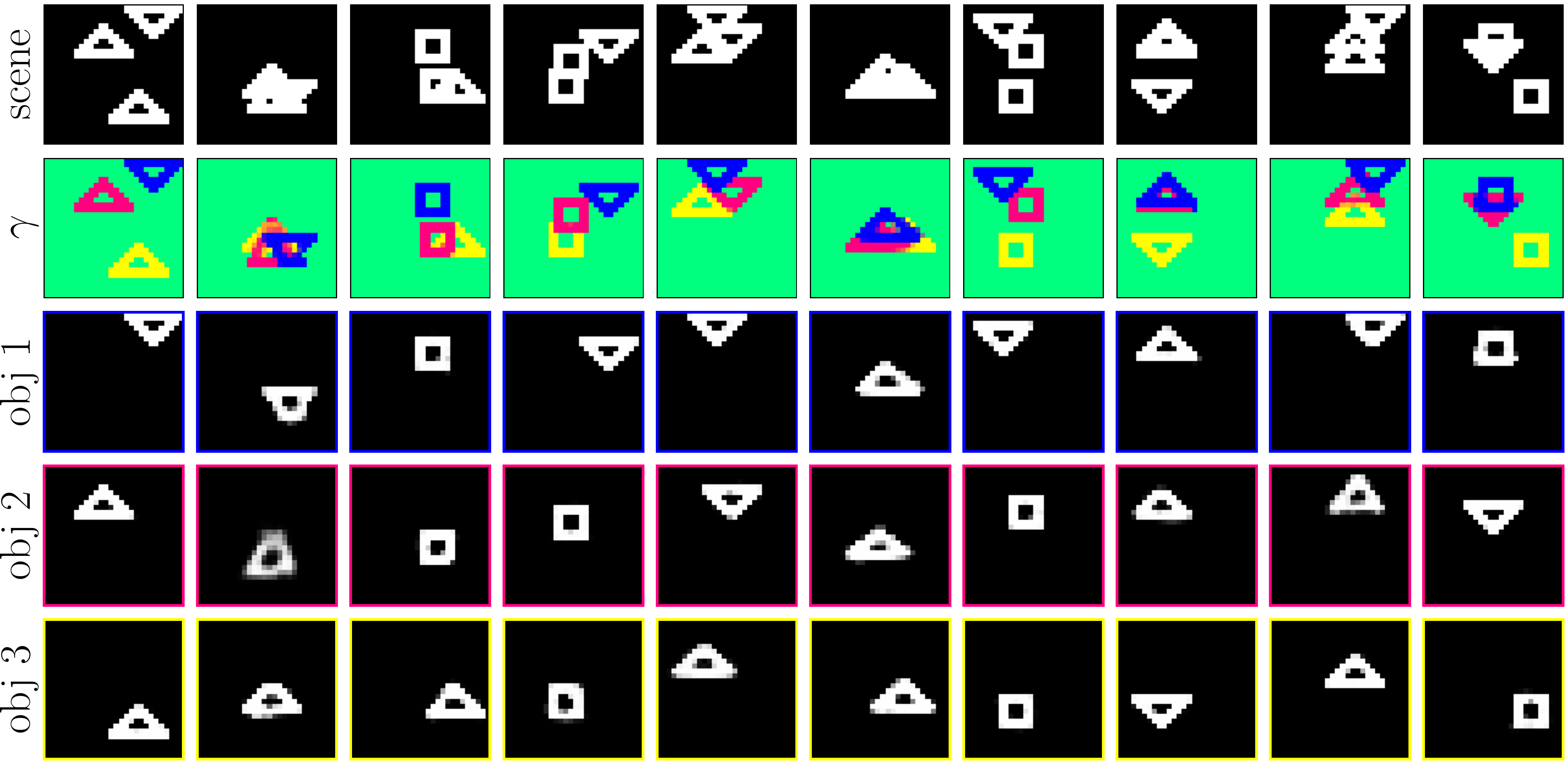}%
		\label{fig:shapes_proposed}}
	\\
	\subfloat[Softmax (Size $28 \times 28$)]{\includegraphics[width=\linewidth]{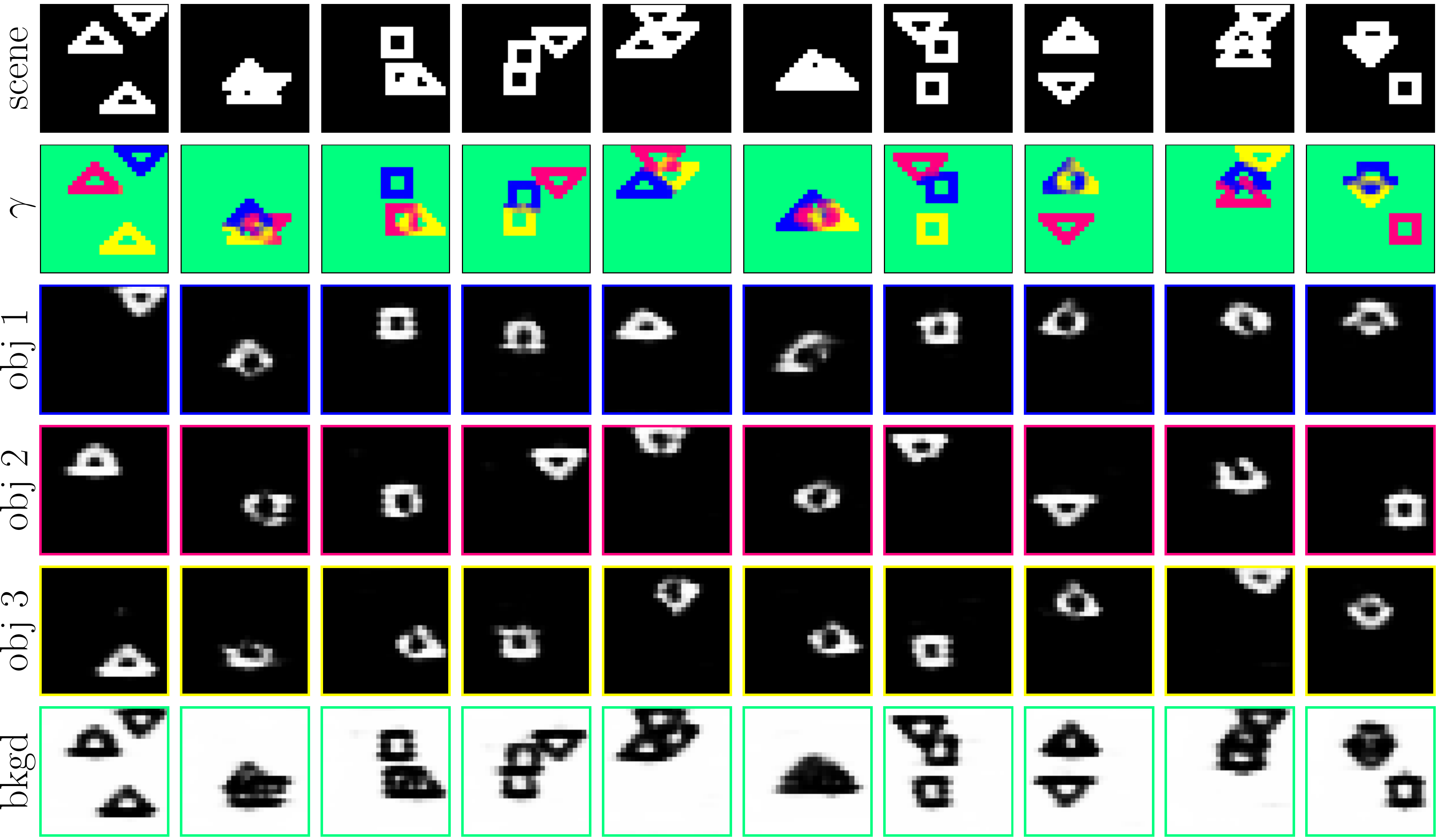}%
		\label{fig:shapes_softmax}}
	\caption{Component assignments and reconstructed images of individual objects evaluated on the Multi-Shapes dataset.}
	\label{fig:shapes_compare_methods}
\end{figure}

Performances of Tagger, N-EM and the proposed method are compared on two subsets of the Multi-Shapes dataset which differ from each other in image size ($20 \!\times\! 20$ and $28 \!\times\! 28$). Images in both subsets contain $3$ objects. Because objects in both subsets are of the same size, $20 \!\times\! 20$ images which consist of less pixels are more likely to contain overlapped objects, and performing perceptual grouping on them is not necessarily easier. For Tagger, parameters of mixture models are updated via a $3$-layer Ladder Network. For N-EM and the proposed method, the encoder and decoder networks are convolutional neural networks (CNNs) with $2$ convolutional, $2$ fully-connected and $3$ layer normalization \cite{ba2016layer} layers. The qualitative results are presented in Figure \ref{fig:shapes_compare_methods}, and the AMI/MSE scores are presented as follows.

\begin{center}
	\begin{tabular}{c C{1in} C{1in}}
		\hline
		& Size $20 \times 20$ & Size $28 \times 28$ \\ \hline
		Tagger & \textbf{0.933} / 0.71e-2 & 0.820 / 1.49e-2 \\
		N-EM & 0.824 / 1.59e-2 & 0.897 / 0.82e-2 \\
		Proposed & 0.920 / \textbf{0.40e-2} & \textbf{0.941} / \textbf{0.28e-2} \\
		\hline
	\end{tabular}
\end{center}

On the subset consisting of $20 \times 20$ images, Tagger achieves the highest grouping accuracy (AMI) and the proposed method best estimates the images of individual objects (MSE). The AMI score of the proposed method is slightly lower than Tagger. On the other subset, The proposed method achieves both the best AMI and MSE scores.

According to Figure \ref{fig:shapes_tagger} (\textit{Tagger}), the learned spatial dependencies of pixels are entangled. Both mixture components (row 2; cols 2--5, 7--10) and mixture weights (row 3; cols 2--5, 7--10) contain partial and imperfect information of shape and appearance. It is caused by modeling both mixture components and mixture weights to be spatially dependent without adding additional regularizations. Another observation is that objects are mainly distinguished based on mixture components, and the mixture weights assist the segregation when ambiguity exists. Although the mixture component assignments are predicted accurately for foreground objects (row3; cols 1, 6), the background pixels are not well inferred based on posterior probabilities $\gamma$ (row2; cols 1, 6). The possible reason is that the diverse combinations of foreground objects prevent background pixels from being modeled well using non-combinational representations.

In Figure \ref{fig:shapes_nem} (\textit{N-EM}), the colors of background pixels in the $2$nd row are gray (red+green+blue) because the posterior probabilities $\gamma$ are almost identical for all mixture components. Similar to Tagger, background pixels cannot be well determined via the Maximum \emph{a posteriori} (MAP) estimation in N-EM. This is because N-EM does not use a mixture component to model the background.

As demonstrated in Figure \ref{fig:shapes_proposed} (\textit{proposed method}), objects are well segregated based on the learned knowledge of spatial dependencies even if objects are heavily overlapped (cols 2, 6). Although the reconstructed images of individual objects are not completely correct, by modeling the background complementary to the foreground objects and disentangling different attributes of objects, the predicted results follow the principles of proximity, similarity, continuity, and closure in the Gestalt laws of grouping.

\subsection{Effectiveness of Modeling Background Separately}

Besides the observation from Figure \ref{fig:shapes_tagger} that modeling the background as an ordinary component results in imperfect separation of foreground and background based on posterior probabilities, we also assess the effectiveness of modeling the background separately by evaluating the performances of the softmax method on the second subset (size $28 \!\times\! 28$) used in the previous experiment. The best AMI and MSE scores achieved are $0.875$ and $1.10 \!\times\! 10^{-2}$, which are significantly worse than the results of the proposed method ($0.941$ and $0.28 \!\times\! 10^{-2}$). The burden of directly representing the complex-shaped background region prevents the model to focus on spatial dependencies of foreground pixels. The qualitative results of the softmax method is shown in Figure \ref{fig:shapes_softmax}. The estimated images of individual objects are not visually plausible although the results of posterior inferences (row 2) are satisfactory.

\subsection{Generalizability and Sensitivity Analyses}

\subsubsection{Generalizability of Compositional Representations}

To illustrate the generalizability of the learned compositional representations, the proposed method are evaluated on three subsets of the Multi-Shapes dataset that differ in the number of objects in images. Results of N-EM are also presented. Models of both methods are trained \textit{only} on the second subset. The AMI/MSE scores are shown below.
\begin{center}
	\begin{tabular}{c C{0.9in} C{0.9in}}
		\hline
		& N-EM & Proposed \\ \hline
		2 Objects & \textit{0.906} / 0.98e-2 & \textit{0.961} / \textit{0.11e-2} \\
		3 Objects (Train) & 0.897 / \textit{0.82e-2} & 0.941 / 0.28e-2 \\
		4 Objects & 0.838 / 1.44e-2 & 0.901 / 0.69e-2 \\
		\hline
	\end{tabular}
\end{center}
For both approaches, the learned representations of conceptual entities generalize well to visual scenes containing more or less objects. The accuracies of estimated mixture component assignment are slightly higher when the task is simpler, and moderately lower when dealing with more challenging visual scenes. The MSE score of N-EM becomes slightly worse on the simpler task.

\subsubsection{Sensitivity to Choice of Conditional Distributions}

\begin{table}
	\centering
	\begin{tabular}{|c|c|c||C{0.37in}|C{0.42in}||C{0.37in}|C{0.42in}|}
		\hline
		& \multirow{2}{*}{$F$} & \multirow{2}{*}{$N$} & \multicolumn{2}{c||}{N-EM} & \multicolumn{2}{c|}{Proposed} \\ \cline{4-7}
		& & & AMI & MSE & AMI & MSE \\ \hline\hline
		\multirow{5}{*}{\STAB{\rotatebox[origin=c]{90}{Bernoulli}}}
		& 16 & 16 & 0.876 & 0.95e-2 & 0.856 & 1.28e-2 \\ \cline{2-7}
		& 32 & 16 & \textit{0.897} & \textit{0.82e-2} & 0.868 & 1.23e-2 \\ \cline{2-7}
		& 32 & 32 & 0.816 & 1.32e-2 & 0.926 & 0.43e-2 \\ \cline{2-7}
		& 64 & 32 & 0.814 & 1.34e-2 & 0.920 & 0.40e-2 \\ \cline{2-7}
		& 64 & 64 & 0.749 & 1.70e-2 & \textbf{0.941} & \textbf{0.28e-2} \\
		\hline\hline
		\multirow{5}{*}{\STAB{\rotatebox[origin=c]{90}{Gaussian}}}
		& 16 & 16 & \textit{0.687} & 2.35e-2 & 0.915 & 0.68e-2 \\ \cline{2-7}
		& 32 & 16 & 0.676 & \textit{2.34e-2} & 0.911 & 0.63e-2 \\ \cline{2-7}
		& 32 & 32 & 0.566 & 2.87e-2 & 0.914 & 0.51e-2 \\ \cline{2-7}
		& 64 & 32 & 0.503 & 3.06e-2 & 0.935 & 0.30e-2 \\ \cline{2-7}
		& 64 & 64 & 0.353 & 3.77e-2 & \textbf{0.939} & \textbf{0.29e-2} \\
		\hline
	\end{tabular}
	\caption{Comparisons of AMI and MSE scores with different mixture component distributions. $F$ and $N$ are respective dimensions of RNN inputs and latent representations.}
	\label{tab:shapes}
\end{table}

We assess the influences of different choices of mixture component distributions on the second subset ($3$ objects) used in the previous experiment. Detailed performances of N-EM and the proposed method are shown in Table \ref{tab:shapes}. If intensities of binary images are modeled by Gaussian distributions, the grouping results of N-EM deteriorate drastically. The proposed method, on the other hand, are less sensitive to the form of mixture component distributions.

\subsubsection{Sensitivity to Diversity of Objects}

\begin{table}
	\centering
	\begin{tabular}{|c|c|c||C{0.37in}|C{0.42in}||C{0.37in}|C{0.42in}|}
		\hline
		& \multirow{2}{*}{$F$} & \multirow{2}{*}{$N$} & \multicolumn{2}{c||}{N-EM} & \multicolumn{2}{c|}{Proposed} \\ \cline{4-7}
		& & & AMI & MSE & AMI & MSE \\ \hline\hline
		\multirow{4}{*}{\#1}
		& 32 & 16 & \textit{0.780} & \textbf{1.17e-2} & 0.748 & 1.89e-2 \\ \cline{2-7}
		& 32 & 32 & 0.573 & 1.54e-2 & 0.765 & 1.64e-2 \\ \cline{2-7}
		& 64 & 32 & 0.521 & 1.54e-2 & 0.787 & 1.35e-2 \\ \cline{2-7}
		& 64 & 64 & 0.391 & 1.79e-2 & \textbf{0.790} & \textit{1.30e-2} \\
		\hline\hline
		\multirow{4}{*}{\#2}
		& 32 & 16 & \textit{0.317} & 2.86e-2 & \textbf{0.719} & 2.07e-2 \\ \cline{2-7}
		& 32 & 32 & 0.254 & 2.75e-2 & 0.704 & 2.04e-2 \\ \cline{2-7}
		& 64 & 32 & 0.194 & \textit{2.71e-2} & 0.719 & \textbf{1.61e-2} \\ \cline{2-7}
		& 64 & 64 & 0.152 & 2.73e-2 & 0.676 & 1.63e-2 \\
		\hline\hline
		\multirow{4}{*}{\#3}
		& 32 & 16 & \textit{0.335} & 2.93e-2 & \textbf{0.715} & 2.16e-2 \\ \cline{2-7}
		& 32 & 32 & 0.223 & 3.15e-2 & 0.702 & 2.07e-2 \\ \cline{2-7}
		& 64 & 32 & 0.185 & 2.94e-2 & 0.705 & 1.69e-2 \\ \cline{2-7}
		& 64 & 64 & 0.158 & \textit{2.84e-2} & 0.683 & \textbf{1.62e-2} \\
		\hline
	\end{tabular}
	\caption{AMI and MSE scores evaluated on different subsets of the Multi-MNIST dataset. $F$ and $N$ are respective dimensions of RNN inputs and latent representations.}
	\label{tab:static_mnist}
\end{table}

The Multi-MNIST dataset is much more challenging than the Multi-Shapes dataset for the larger image size ($48 \!\times\! 48$ versus $20 \!\times\! 20$ or $28 \!\times\! 28$), as well as higher degrees of both inter-object variations (the number of unique objects) and intra-object variations (intensities of pixels belonging to the same object). There are $3$ subsets in this dataset. Subsets 1, 2 and 3 are constructed using $20$, $500$ and $\num[group-separator={,}]{70000}$ unique digits, respectively. Different from the first two subsets, the digits composing images in the test set of subset 3 do not exist in the training and validation set. We evaluate the performances of N-EM and the proposed method on these subsets, and compare the sensitivities to object diversities of the two approaches. Experimental results are presented in Table \ref{tab:static_mnist}.

When the diversity of objects is small, N-EM and the proposed method perform similarly well. N-EM achieves slightly higher AMI score and the proposed method attains lower MSE score. As the variations of objects increase, the performance of the proposed method drops moderately, and the grouping results of N-EM decline significantly. On subsets 2 and 3, both methods achieve the highest AMI scores when dimensions of RNN inputs and latent representations are small, and the best MSE scores when these dimensions are slighly larger. The possible reason is that higher dimensions allow models to focus on more details and help reconstruct individual digits. The large diversities of objects, on the other hand, misguide models to represent parts of different objects in the same mixture component and lower the grouping accuracies as the capacities of models increase.

\begin{figure}[t]
	\centering
	\subfloat[N-EM (20 variants)]{\includegraphics[width=0.5\linewidth]{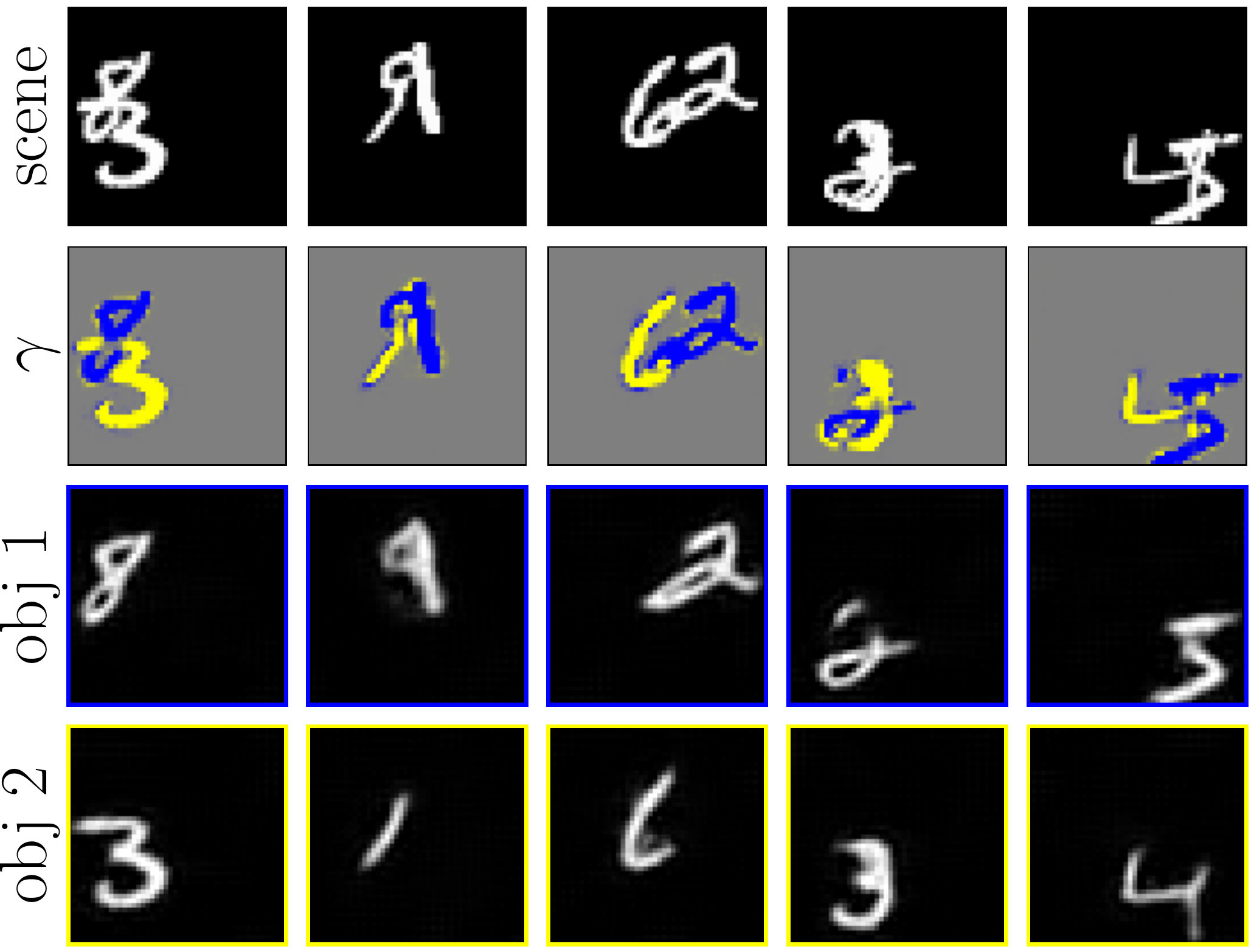}%
		\label{fig:mnist_20_nem}}
	\hfil
	\subfloat[Proposed (20 variants)]{\includegraphics[width=0.5\linewidth]{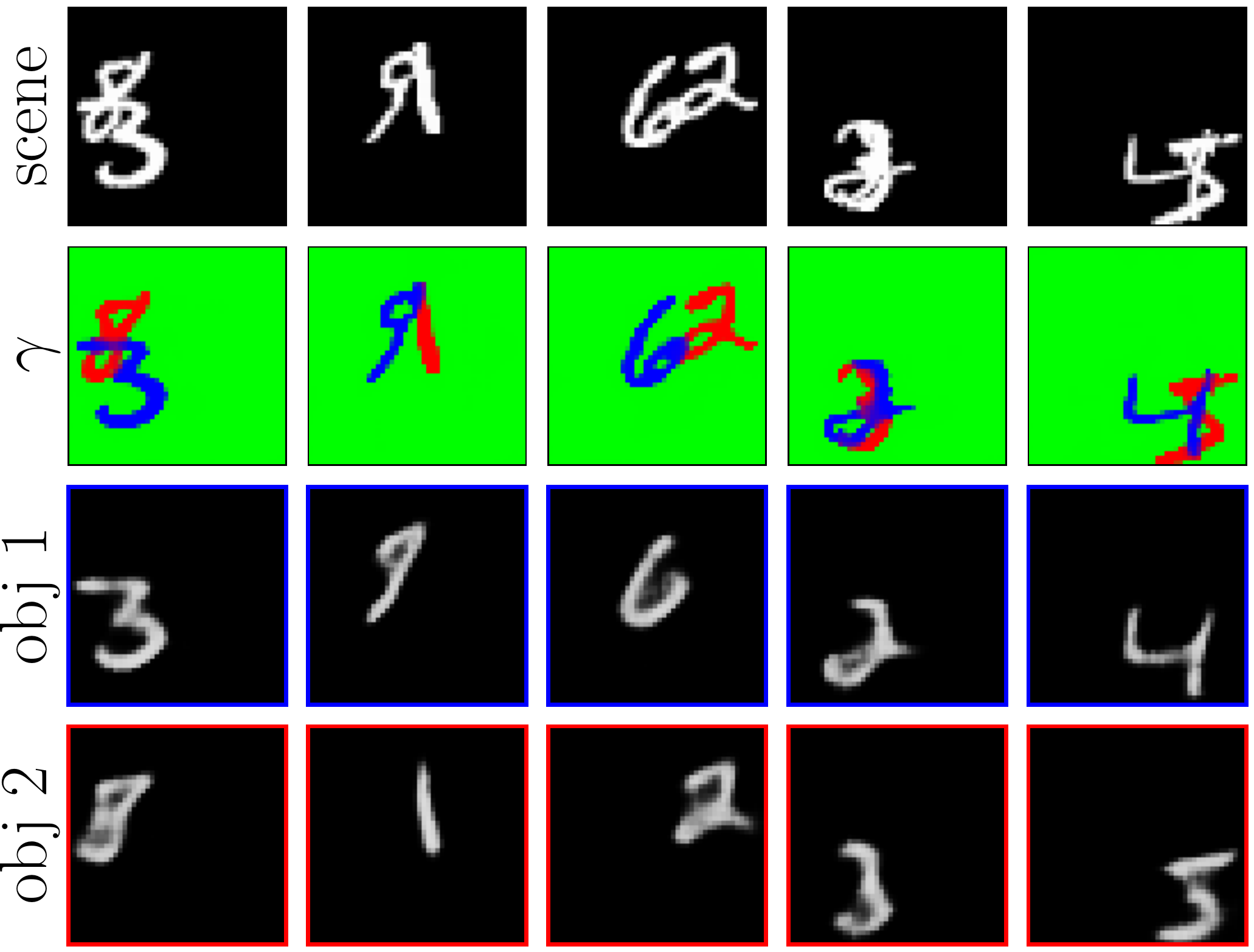}%
		\label{fig:mnist_20_proposed}}
	\\
	\subfloat[N-EM (500 variants)]{\includegraphics[width=0.5\linewidth]{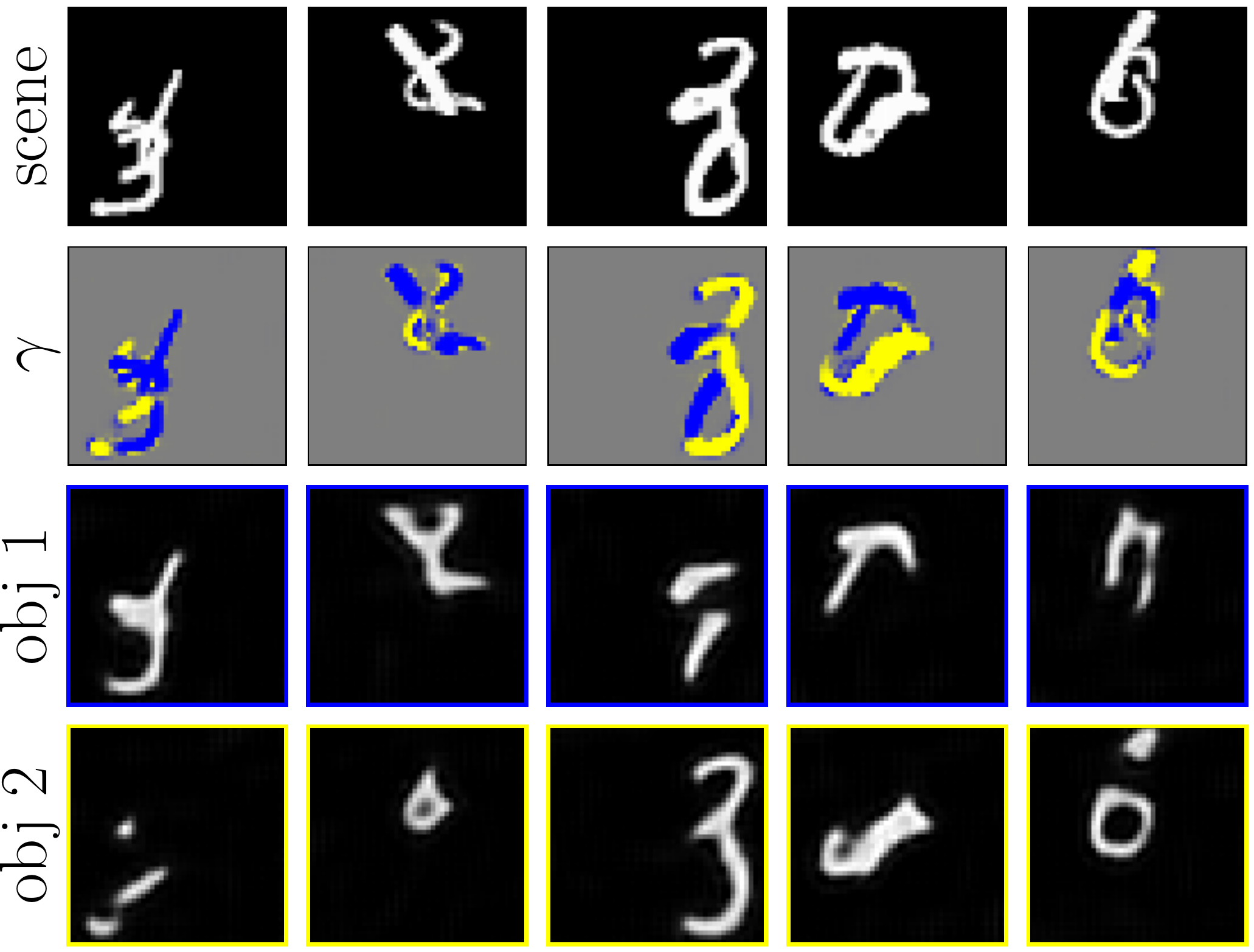}%
		\label{fig:mnist_500_nem}}
	\hfil
	\subfloat[Proposed (500 variants)]{\includegraphics[width=0.5\linewidth]{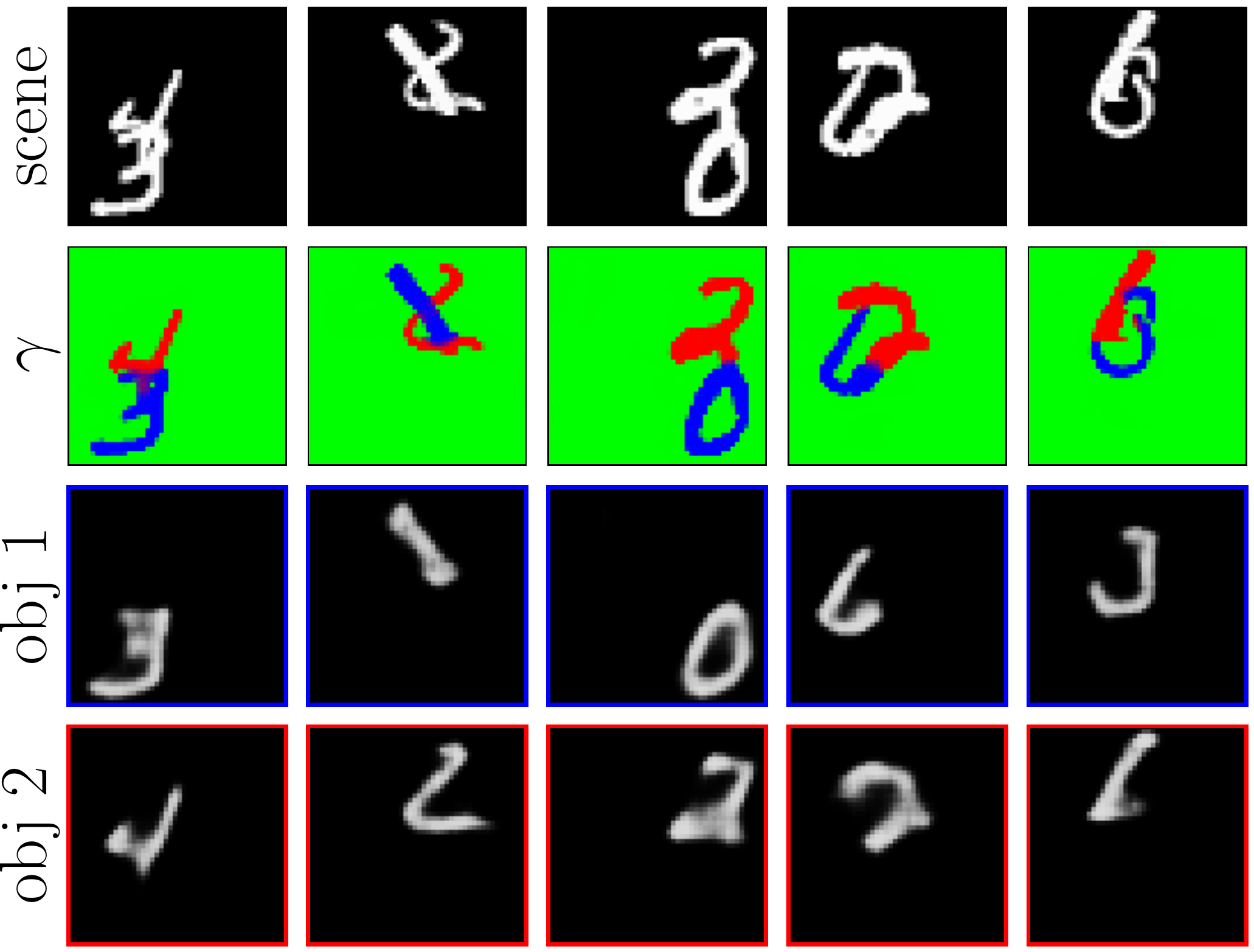}%
		\label{fig:mnist_500_proposed}}
	\\
	\subfloat[N-EM (70,000 variants)]{\includegraphics[width=0.5\linewidth]{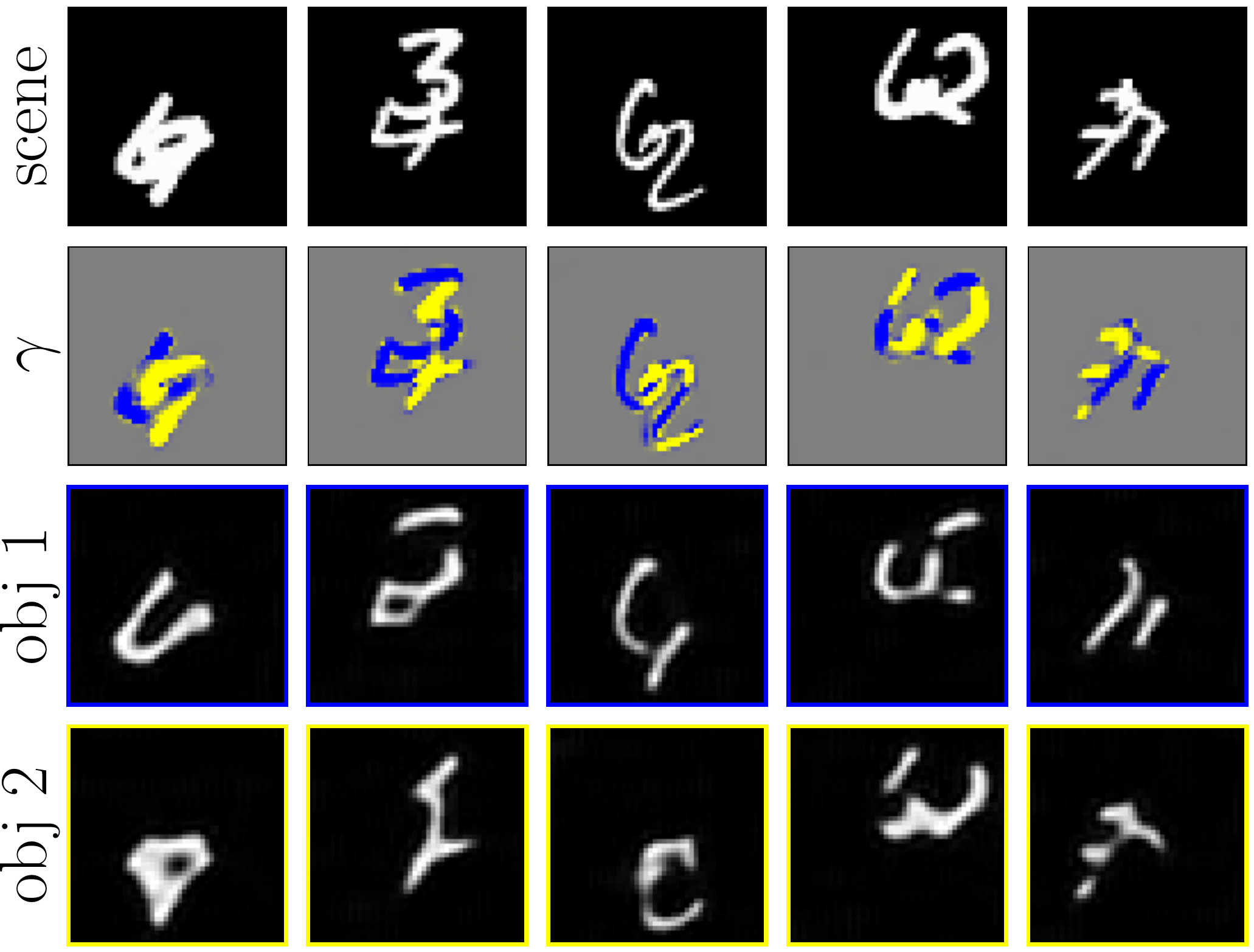}%
		\label{fig:mnist_all_nem}}
	\hfil
	\subfloat[Proposed (70,000 variants)]{\includegraphics[width=0.5\linewidth]{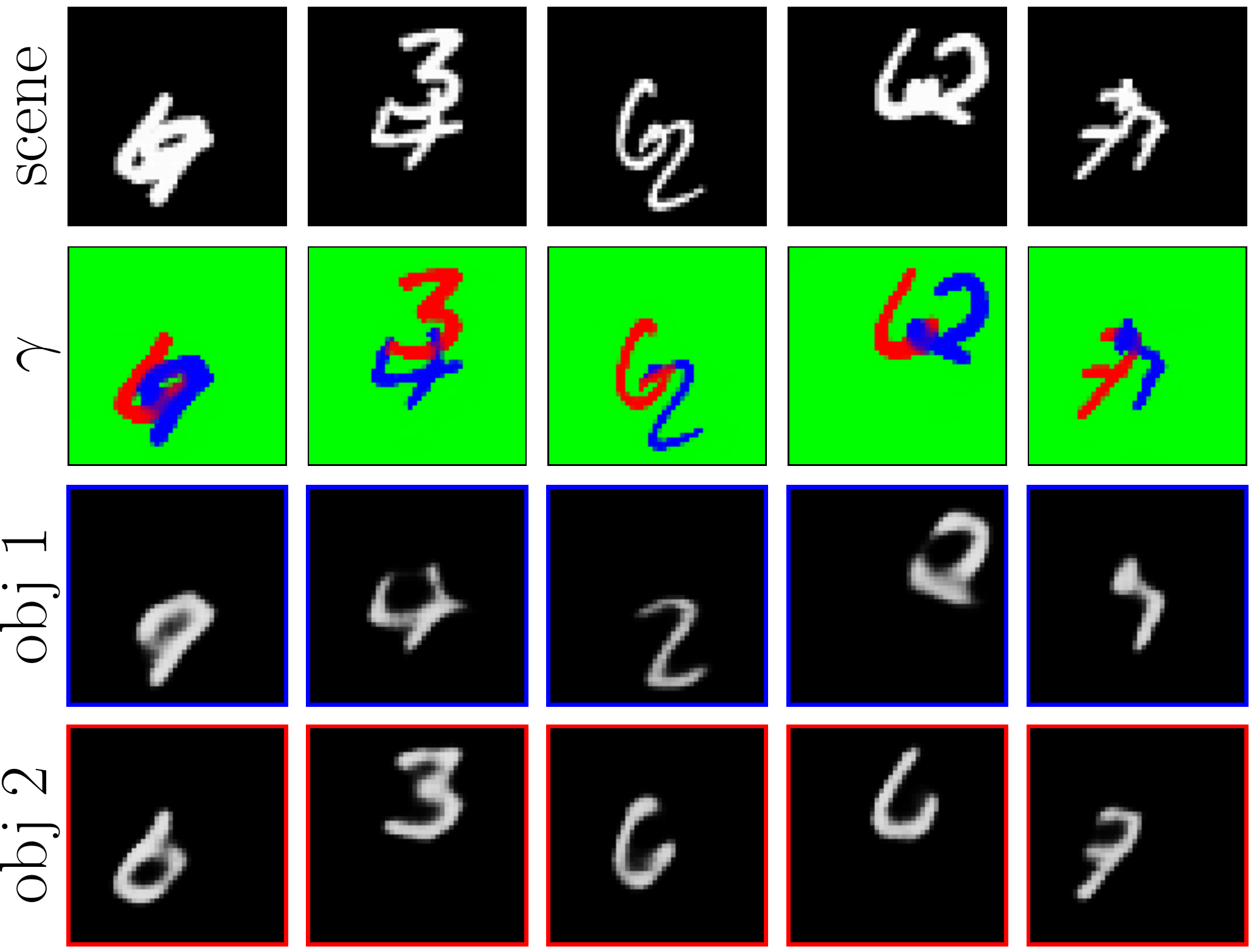}%
		\label{fig:mnist_all_proposed}}
	\caption{Component assignments and reconstructed images of individual objects evaluated on the Multi-MNIST dataset.}
	\label{fig:mnist}
\end{figure}

Figure \ref{fig:mnist} demonstrates the predicted component assignments and reconstructed images of individual objects produced by both methods. On subset 1, the results generated by N-EM are slightly more visually plausible. Except for the occluded regions, details of each object are accurately reconstructed. This is because N-EM allows the mixture components to vary among pixels, which helps better reconstruct the intra-object variations of each digit. Both methods can accurately segregate objects from some complex visual scenes like the last two images in Figure \ref{fig:mnist_20_nem} and \ref{fig:mnist_20_proposed}. On subsets 2 and 3, the estimated images of individual objects generated by N-EM are also sharper than the proposed method. However, each component contains pixels of multiple objects. The proposed method models the spatial dependencies of pixels with mixture weights, and utilizes the principle of similarity implied by the spatially independent mixture components to assist the learning of correlations between pixels. Empirical results suggest that this scheme is less sensitive to the diversities of objects in the visual scenes.

\section{Conclusion}

In this paper, we have proposed a framework called Learnable Deep Priors (LDP) which defines priors of model parameters with compositional latent representations and learnable neural networks for tackling the perceptual grouping problem. By modeling the background as a special component in a compositional manner and decomposing the attributes of conceptual entities into shapes and appearances that are separately represented, the proposed method better learns spatial dependencies than existing methods for perceptual grouping. We have demonstrated that the proposed method is insensitive to diversities of objects, and the learned compositional representations of individual entities generalize well to visual scenes constructed by novel combinations of these entities. Further research in this direction could be considering learnable deep priors with latent representations of conceptual entities organized in a structured form, integrating hierarchical spatial mixture models with neural networks, and solving higher-level tasks like relational inferences based on the learned compositional representations.

\section{Acknowledgments}

This work was supported in part by National Key R\&D Program of China (No.2017YFC0803700), NSFC under Grant No.61572138 \& No.U1611461, and STCSM Projects under Grant No.16JC1420400 \& No.18511103104.

\end{document}